\documentclass[5p,twocolumn,preprint]{elsarticle}  
\usepackage{epsfig} 
\usepackage{url}
\usepackage{rotating}

\usepackage{amsmath} 
\usepackage{amssymb}  
\usepackage{multirow}

\usepackage{multicol}        
\usepackage[bottom]{footmisc}
\usepackage{algorithm}
\usepackage{algorithmic}
\usepackage{epstopdf}
\usepackage{subfig}
\usepackage{xcolor}

\usepackage{hyperref}

\usepackage[switch]{lineno}

\graphicspath{{figures/}}
\begin{document}


\begin{frontmatter}
\title{\LARGE \bf A Survey of Knowledge Representation in Service Robotics}

\author{David Paulius}
\ead{davidpaulius@usf.edu}
\cortext[cor1]{Corresponding author}
\author{Yu Sun\corref{cor1}}
\ead{yusun@usf.edu}
\address{Department of Computer Sci \& Eng, University of South Florida, 4202 E. Fowler Ave, Tampa, FL, United States, 33620}

\begin{abstract}
Within the realm of service robotics, researchers have placed a great amount of effort into learning, understanding, and representing motions as manipulations for task execution by robots.
The task of robot learning and problem-solving is very broad, as it integrates a variety of tasks such as object detection, activity recognition, task/motion planning, localization, knowledge representation and retrieval, and the intertwining of perception/vision and machine learning techniques.
In this paper, we solely focus on knowledge representations and notably how knowledge is typically gathered, represented, and reproduced to solve problems as done by researchers in the past decades.
In accordance with the definition of knowledge representations, we discuss the key distinction between such representations and useful learning models that have extensively been introduced and studied in recent years, such as machine learning, deep learning, probabilistic modelling, and semantic graphical structures.
Along with an overview of such tools, we discuss the problems which have existed in robot learning and how they have been built and used as solutions, technologies or developments (if any) which have contributed to solving them.
Finally, we discuss key principles that should be considered when designing an effective knowledge representation.
\end{abstract}


\end{frontmatter}

\section{Introduction}
A recent trend in robot learning research has emerged from the motivation of using robots in human-centered environments to develop domestic robots or robot assistants for use in homes and to automate certain processes and tasks which may be inconvenient for us as humans to perform.
Assistants are particularly beneficial for the care of the elderly or disabled who cannot perform the actions themselves.
In order to develop such robots, researchers aim to create robots which can {\it learn}, {\it understand}, and {\it execute tasks} as human beings would.
Individually, these are very complicated tasks, as they all require an introspective approach to understanding: 1) how we would perceive our environment (objects, obstacles, and navigation), 2) how we execute actions, 3) how these actions are best designed and implemented, 4) how familiar we are to objects of various types and attributes, and 5) how we ground  understanding of the world.
To achieve these goals, an integrated solution is needed which allows us to perform a variety of tasks such as object detection, activity recognition, task/motion planning, localization, knowledge representation and retrieval, and the intertwining of computer vision and machine learning.
Furthermore, aside from handling variability, such a solution must ensure that robots avoid harming any humans 
around them and that they maintain a safe environment through the understanding of the intentions of humans and the effects of its own actions.
In this paper, we pay special attention to the way knowledge is represented for retention and re-use; a \emph{knowledge representation} can make it easier for a robot to perform its duties and to function in its workspace safely.

\subsection{Defining Knowledge Representation}
\label{sec:define}
\begin{figure}[t]
	\centering
	\includegraphics[width=4.5cm]{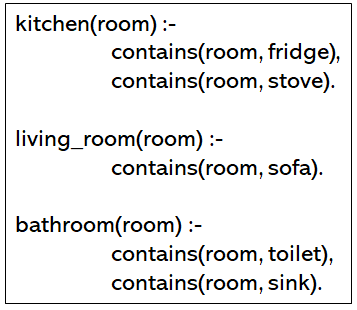}
	\caption{Logical expressions as used in AI and certain robotics applications can be written in different ways such as the Prolog logic programming language. 
	    Here, we represent different predicates that a robot can use to determine whether the room it is in is a kitchen, living room or bathroom based on what is found within it.
	}
	\label{fig:Prolog}
\end{figure}

As established before, several models or tools available at the researcher's disposal have been adopted for learning manipulations or activities to some regard for service or domestic robots.
Although this may be true, only a few actually make the distinction of being a {\it knowledge representation}, and these full-fledged representations cannot be considered equivalent to stand-alone models, tools or classifiers.
However, in our field of robotics, there is a lack of a formal definition of what constitutes a knowledge representation for a robotic system.
Originally, this concept was derived from the field of artificial intelligence (AI), as representation of knowledge is very crucial to building an artificially intelligent agent or system.
As originally defined in AI, a knowledge representation is \emph {``concerned with how knowledge can be represented symbolically and manipulated in an automated way by reasoning programs"} \citep{Brachman03knowledgerepresentation}. 
The emphasis in AI is heavily placed on the semantic and symbolic representation of knowledge grounded in formal logical expressions.
Concepts about the robot's world, its actions, and the consequences of said actions can be described using causal rules and predicates that are connected to one another based on common entities or variables.
By identifying instances of such entities and variables, a robot can make propositions to determine whether or not these expressions are true or false through reasoning.
An agent (such as a robot) can use these expressions as their source of knowledge so long as they can effectively acquire input and use reasoning to determine the correct outcome.
Typically, programmers can take advantage of languages such as Lisp and Prolog to write relations (facts or rules) as clauses that can be used in queries.
Figure \ref{fig:Prolog} shows an example of how logical expressions can be formulated in the Prolog logic programming language.
Within such knowledge representations, rules or concepts that are built into the agent are referred to as explicit knowledge, and this knowledge is used to infer new concepts which were initially unknown to the agent as implicit knowledge.

Ideally, a robot's knowledge representation combines several forms of knowledge with reasoning capabilities such that it can devise the optimal plan of action for performing its duties.
In robotics, a knowledge representation extends beyond AI's logical formalism of knowledge using only predicates to other components that are important to a robotic system: perception modules (audio, vision, or touch), sensor inputs, motion primitives and actuation systems, motion and task planning systems, inference systems, et cetera.
As an extension to the definition in AI, we define a knowledge representation for robotics as \emph{a means of representing knowledge about a robot's actions and environment, as well as relating the semantics of these concepts to its own internal components, for problem solving through reasoning and inference.
}
Here, high-level knowledge denotes a semantic, structural representation of relations between different components, while low-level representations on the other hand have no structure nor symbols and instead relate to the robot's control program or physical system.
For instance, we would typically describe objects and locations using prepositions or very descriptive grammar, which is unlike the low-level parameters used by robots such as sensor readings, joint angles or points in space.
Simply speaking, a knowledge representation gives {\it meaning} to the inputs that a robot acquires and uses in its tasks.
For instance, a robot can follow a certain trajectory or primitive to perform a task, but high-level knowledge can be used to give meaning to it, such as defining what the manipulation is or does.
This problem relates to the {\it symbol grounding problem }\citep{harnad1990symbol}, which is concerned with understanding how to intrinsically draw meaning (as high-level) to symbols (as low-level) through experience and interaction with the world; a knowledge representation for a robot or artificially intelligent agent has to answer the question of meaning behind what it may see or do in its environment.
This symbol grounding problem states that we cannot merely associate symbols to concepts or meaning without considering how these connections are established by the intelligent agent.
This along with the problem of symbol grounding is covered extensively in \citep{taniguchi2016symbol}.
For human-like manipulations and task planning within our homes and surroundings, if a robot were to communicate with a human user and understand commands and relations as dictated by a human person, then the knowledge representation must have an ambivalent description of objects/tools and manipulations in high- and low-level representations. 
Furthermore, a robot should also understand similarly to how we as humans understand the world.
With the inclusion of logical expressions, knowledge representations still can benefit from the powerful capability of inference and deduction as they were intended for use in AI.
Several representations discussed in this paper, for example, ground knowledge in Prolog to make querying relations easier using its programming syntax.

\subsection{Overview of Paper}
This paper aims to discuss the recent trend in knowledge representation for robots, to identify issues in creating effective representations, and to review several tools and models that have been successfully applied to smaller sub-problems in robot learning and manipulation to thus create representations.
We discuss key characteristics of representations that allow them to function in spite of the highly variable nature of a robot's working environment and the objects found within it.
The main contribution of our paper is a focused discussion on what a knowledge representation is as it pertains to robotics: a comprehensive tool that encompasses high-level knowledge and low-level features. 
We also describe how models can be used to perform specific tasks that are key to a robot's function.

Our paper is outlined as follows:
\begin{enumerate}[$\bullet$]
\item{
Firstly, we delve into a discussion on exemplary high-level knowledge representations that are comprehensive and complete to be called such in Section \ref{sec:com}. 
Section \ref{sec:com} gives an overview of those knowledge representations, with special attention paid to cloud-based knowledge representations and cognitive architectures.
}
\item{
Next, we examine numerous specialized knowledge representations in Section \ref{sec:models}.
These models have been effective in representing properties and concepts that are crucial to activity understanding and task execution, despite that they represent very specific concepts.
}
\item{
In Section \ref{sec:ML}, we briefly talk about classifiers designed for important tasks in robotics, despite not explicitly describing rules and semantics for understanding.
}
\item{Finally, in Section \ref{sec:eval}, after reviewing several ways of representing knowledge, we address key issues in developing knowledge representations.
Specifically, we propose key concerns that should be addressed to build an effective representation while relating them to discussed works.}
\end{enumerate}

\section{Comprehensive Knowledge Representations}
\label{sec:com}

In the ideal world, a domestic robot would be directed by commands as input and it should plan its actions and execute a series of tasks to produce a required outcome.
To do this, the robot should be equipped with the knowledge it needs to perform all the steps, from receiving its instructions to identifying a motion plan to executing all required manipulations using what is available in its environment.
We first begin our discussion on knowledge representations by introducing examples of those that have been introduced by researchers within recent years.
These representations should contain an abstraction of low-level features (features that can be acquired by a robotic system) with high-level knowledge that can be interpreted by humans.
Knowledge can be constrained to a set of terms or language known as an {\it ontology}.
The purpose of an ontology is to define a scope of concepts and terms used to label and describe the working space of the robot in a format which is also understood by humans.

In this section, we will discuss different approaches to mapping a higher level of knowledge to a robot's understanding of the world to perform tasks.
These representations combine several kinds of information.
In addition, we discuss larger comprehensive projects that emphasize collaboration and knowledge sharing among a group of robotic systems.
These representations are self-contained systems which combine multiple modalities of information such as vision, touch and sound, or they can draw information from several data sets all combine and contribute modality information.
A good overview of data sets that offer several modalities of information to researchers related to object manipulation can be found in \citep{huang2016recent}.
This would imply that robots should maintain a constant connection to access past experiences of robots as well as to upload their current and future experiences to build upon that informational source.
Other systems in this section also automatically gather data from across the Web.

\subsection{High-level Representations of Tasks}
\label{sec:high-level}
A knowledge representation innately represents skills or actions in an ontology, emphasizing reliability, safety and usability for and by robots.
In order for the robot to perform effectively, a knowledge representation should contain concepts or rules based on different modalities of data, it should allow for expansion or learning new concepts, it should be used to reason logically and statistically, and that it appropriately defines objects, actions/skills, and states needed for manipulations.
As we will address in more detail in Section \ref{sec:eval}, a knowledge representation should encompass all of these ideas so that a robot can reason as well as perform skills on its own.
When representing skills, it is not only important to consider motion-related details such as trajectories, but we should also  consider the semantic meaning behind the skills (i.e. what exactly is the motion, what changes or consequences do its actions cause on the robot's environment, etc.).
Many researchers do not consider the representation of states within their work \citep{jelodar2018identifying}, which is important for recognizing when an action is complete.
We pay particular attention to knowledge representations that ground tasks as high-level knowledge through the integration of different structures and models.

We discuss our first example of a high-level knowledge representation, which was originally proposed by Ramirez-Amaro et al. \citep{ramirez2014bootstrapping,ramirez2015understanding,ramirez2017transferring}.
These researchers used learning by demonstration to teach robots about manipulations obtained directly from demonstrations, and they describe it as a transfer of skills from the demonstrator to the robot; upon observation of a demonstration of a skill, the robot then imitates the action performed by a human demonstrator.
This sense of "transfer learning" however is different to the traditional sense within the machine learning community \citep{pan2010survey}.
They can create semantic graphs as trees with knowledge in the form of transferable skills needed to execute three challenging kitchen tasks.
This knowledge is directly extracted from human demonstrators and it allows the robot to perform the exact methods needed to imitate the demonstrator in manipulating objects.
Human activities are learned based on several attributes: 1) the motion made by the hand, 2) the object(s) being moved and manipulated by the hand, and 3) the object(s) which these held items are being used and acted on, and they are presented as ordered pairs to train their inference engine.
Once obtained, these semantic rules, grounded in Prolog, can be used in reasoning and future understanding of demonstrations through inference; these properties were applied to a decision tree classifier to automatically gain knowledge and rules from new demonstrations.

\begin{figure}[t]
	\centering
	\includegraphics[width=0.9\columnwidth]{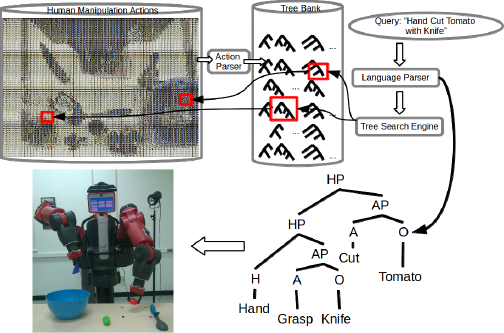}
	\caption{Illustration of the pipeline for manipulation action tree banks as proposed in \citep{yang2014manipulation}. The action tree bank comprises of knowledge from demonstrations, and this knowledge base is searched for a tree which a robot can use for manipulation execution.}
	\label{fig:MAB}
\end{figure}

Another exemplary study in representing skills from demonstrations was presented by Yang et al.
In \citep{YangAFA15}, the researchers presented a method for representing observed manipulations in the form of combinatory categorial grammar (CCG) using rules originally described in \citep{yang2014cognitive}.
This grammar vividly describes a specific action, the items being used, as well as the consequence of performing such an action \citep{Yang_2013_CVPR}, and each action can also be effectively broken down into smaller sub-actions or sub-activities.
Using said context-free grammars, they also developed {\it manipulation action tree banks} \citep{yang2014manipulation} to represent action sequences as tree data structures that can be executed by a robot.
Equipped with action tree banks, a robot can use the knowledge gathered from multiple demonstrations to determine the actions it needs to take, in the form of a tree, for a given manipulation problem (illustrated in Figure \ref{fig:MAB}).

\begin{figure}[t]
	\centering
	\includegraphics[width=0.9\columnwidth]{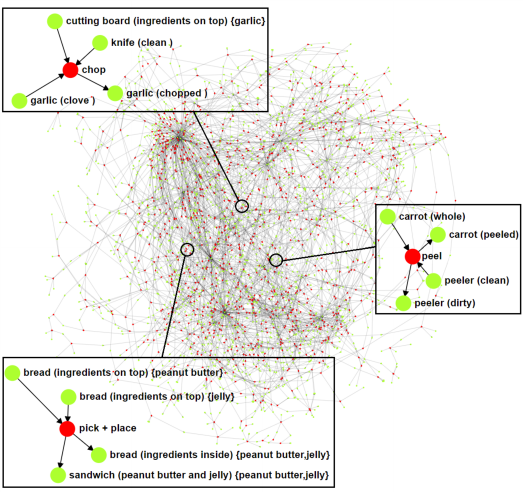}
	\caption{Illustration of a universal FOON as a result of combining knowledge by merging subgraphs from 65 videos. Three examples of functional units are shown, each describing an atomic manipulation.
    }
	\label{fig:FOON}
\end{figure}

Similar to the prior representation, our group also introduced a semantic knowledge representation called the {\it functional object-oriented network} \citep{Paulius2016,Paulius2018,jelodar2018long} (FOON) which serves as a basis for both learning and representing manipulation knowledge for domestic tasks.
The bipartite FOON structure, akin to Petri Nets \citep{Petri:2008}, contains object and motion nodes to capture the concept of object affordances.
Affordances (based on J. J. Gibson's theory of affordance \citep{Gibson_1977}) are described by edges drawn between an object and a motion/action type.
This network combines input object nodes, output object nodes and a motion node to create what we refer to as {\it functional units} (shown in Figure \ref{fig:FOON}).
Individually, a functional unit describes a singular or atomic action that results in an object's change of state before and after a motion is executed upon them (so as to emphasize state change as an indicator of the end of an action \citep{jelodar2018identifying}); collectively, these units describe an entire activity as a subgraph.
Typically, a {\it universal} FOON will comprise of many functional units (shown in Figure \ref{fig:FOON}) obtained from annotating several instructional videos.
Through task tree retrieval (based on graph searching concepts), a robot can find a task tree which contains a sequence of steps that achieve a given goal.
This is done by identifying a target node and backtracking to find steps whose requirements we satisfy (i.e. we have all the objects needed to perform actions), much like the ``firing" of transitions in Petri Nets.
For visualizations of annotated subgraphs, we refer readers to our website\footnote{FOON Website - \url{http://www.foonets.com}}.
To create new FOON subgraphs, in \citep{jelodar2018long}, we proposed a pipeline to annotate videos using a universal FOON.
This deep learning pipeline is capable of suggesting objects and motions as functional units that match activities as seen in new video demonstrations.
The annotations that may be obtained from this process can be merged with our current network to add new knowledge semi-automatically.

\subsection{Cognitive Architectures}
\label{sec:cog}

A {\it cognitive architecture} is a model that attempts to explain the psychological theories behind human cognition and the processes, mechanisms and modules that are pivotal to cognition and behaviour.
Many architectures have been proposed and extensively studied based on different cognitive theories, including the likes of ACT-R \citep{anderson2014rules}, Soar \citep{laird1987soar,laird2012soar}, CLARION \citep{sun2001implicit}, and EPIC \citep{kieras1997overview}.
Every architecture differs from one another in how knowledge is acquired, represented, retained (through long- and short-term memories), and transmitted within its internal cognitive structures since they all follow their own respective theories of human cognition; despite their differences, however, they all aim to answer the question of how human cognition works.
Much like knowledge representations for robots, cognitive architectures modularize different components that are central to thoughts, perception, inference, and action and have them connected with one another based on cognitive theories.
For an extensive overview of such architectures, we encourage readers who are interested in learning more about them to refer to a review done by Kotseruba and Tsotsos \citep{kotseruba2016review}.
Within robotics, cognitive architectures are extensively studied in the field of developmental robotics as a means of understanding how we develop sensory and motor skills as an infant.
Such studies look at how each internal component is connected with one another so that a robot can develop new skills and acquire knowledge about its problem domain.
Readers can learn more about developmental robotics through \citep{lungarella2003developmental,asada2009cognitive,min2016affordance}.

Typically, cognitive architectures emphasize on how knowledge is retained and used as either long- or short-term memory, where long-term memory can refer to concepts (goals or descriptions of an entity's environment), while short-term memory refers to instances of such concepts.
Each of these concepts and skills are learned, retained, and activated once their arguments have been fulfilled by identifying them through the robot's perception system.
Architectures such as Soar \citep{laird2012cognitive} and ICARUS \citep{langley2006unified,kim2010autonomous,kim2011controlling} have been used to illustrate how skills and concepts are learned from navigating throughout its environment or through human interaction. 
They can learn concepts such as object instance identification and location.
These skills (i.e. its abilities to manipulate objects) are learned as long-term components which can be used in conjunction with other skills for performing tasks on the robot's environment.
Through a developmental approach, works such as \citep{vernon2007icub,metta2010icub,ivaldi2012perception,trafton2013act} investigate how skills are developed through human-robot interaction (HRI).
In such studies, a human assistant would interact with robots to teach them about the relationships between its actions, effects, and its own internal representation of the world.
For example, in Ivaldi et al. \citep{ivaldi2012perception}, a robot was taught about object affordance through the assistance of a caregiver who supervises the robot's actions much like a parent would.
The caregiver can give the robot different commands such as looking, grasping, pushing or more complex actions like placing objects on top of another. 
Through HRI, the robot learns how to identify novel objects by showing the robot what the items of focus are without any prior knowledge of what they look like.
Once the robot has the knowledge of those objects, the robot can proceed to learn about actions and the action's effects while gaining positive or negative feedback to indicate whether the robot performed the task correctly or not.
In summary, cognitive architectures not only aim to equip robots with the necessary knowledge to perform its duties, but they ultimately aim to explore how our cognition as humans work.
Retaining a memory of experiences is crucial to learn semantic and symbolic concepts and to reason for solving problems. 

\subsection{Distributive and Collaborative Representations}
\label{sec:cloud}

There is major potential to improve a service robot's performance using cloud computing \citep{remy2011distributed,guizzo2011robots,kehoe2015survey}.
Thanks to cloud computing and powerful hardware, it has become easier for us to perform high-intensive processing tasks with parallel programming and distributed systems.
A good overview of cloud robotics and relevant technologies can be found at \citep{kehoe2015survey}.
Robots can take advantage of off-board computations to allow for longer usage of simpler robotic systems and to reduce their computing power, thus maximizing the overall performance of robots.
Furthermore, certain initiatives such as the Million Object Challenge \citep{oberlinacquiring} and \citep{levine2016learning} show that we make it easier to learn certain tasks in parallel with many robots performing tasks at once.
Highlighted as an important, emerging technology \citep{schaffer_2016}, the Million Object Challenge\footnote{\url{http://h2r.cs.brown.edu/million-object-challenge/}} involves learning how to grasp using Baxter robots working in a distributed manner.
In \citep{levine2016learning}, robots can learn how to grasp using a simple monocular lens camera; 
using such a basic vision system can be applied to other systems to simplify grasping by using more widely accessible and less complicated technologies.
These projects aim to learn how to grasp objects of different shapes and sizes by tasking a group of robots to repeatedly try to pick up a set of objects until they have successfully grasped and placed it in another container.
The important thing to note here is that the \emph{distribution of tasks} can greatly benefit robot learning.

\begin{figure}[t]
	\centering
	\includegraphics[width=0.9\columnwidth]{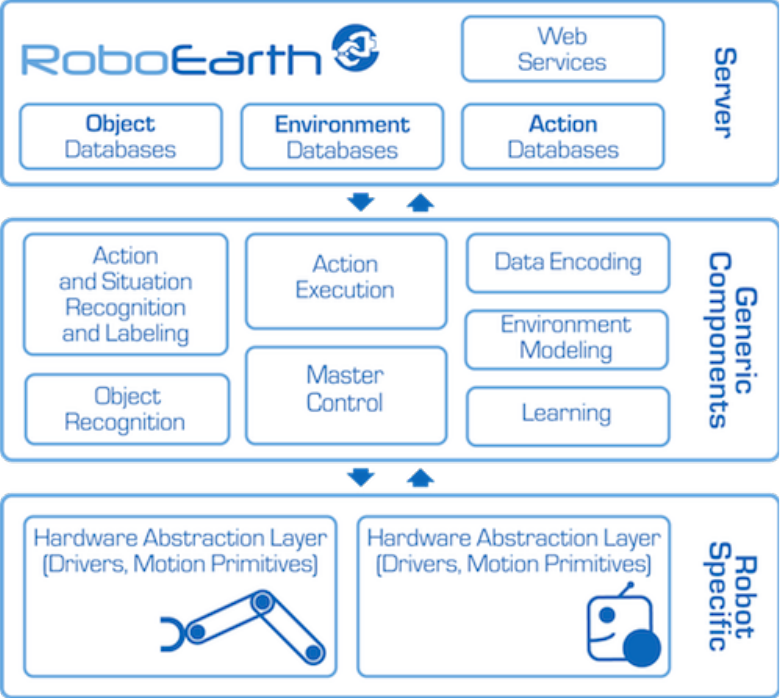}
	\caption{Illustration of the pipeline for RoboEarth \citep{waibel2011roboearth,riazuelo2015roboearth} (as taken from {\url{http://www.i6.in.tum.de/Main/ResearchRoboEarth/}}).
	}
	\label{fig:knowrob}
\end{figure}

\begin{figure*}[t]
	\centering
	\includegraphics[width=15cm]{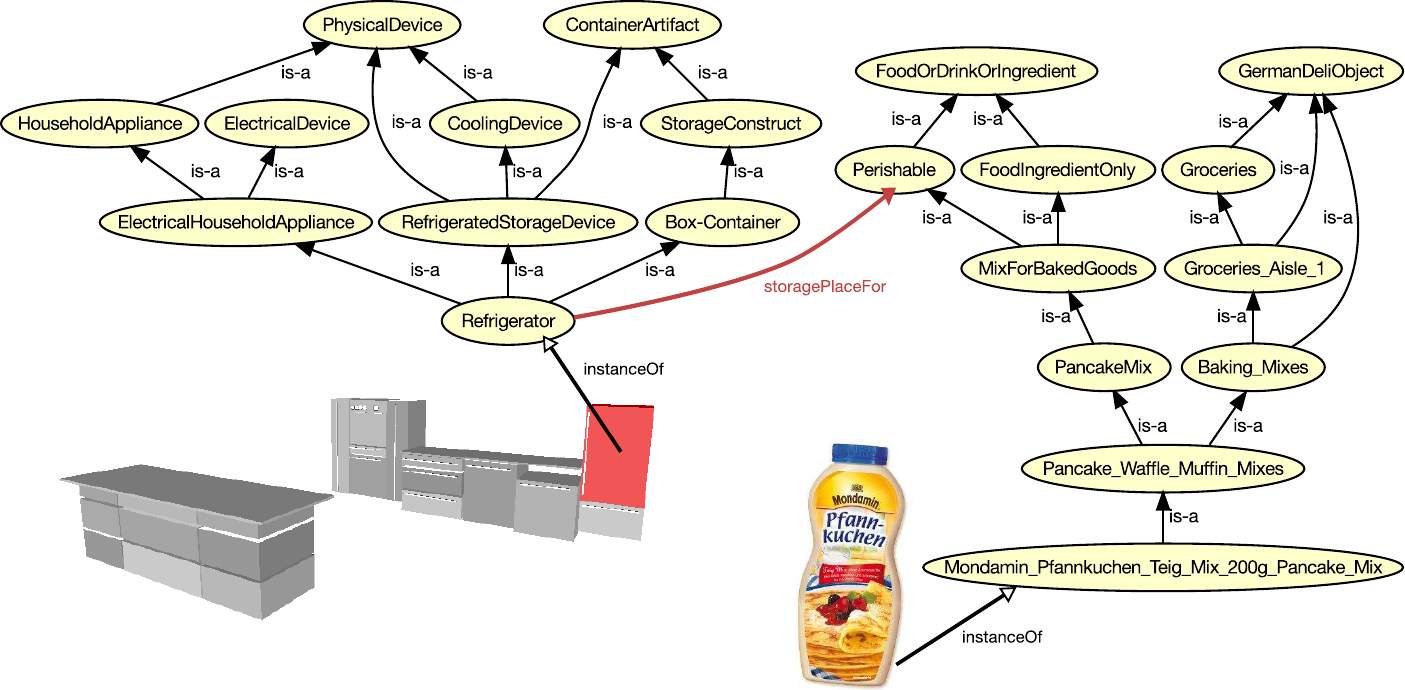}
	\caption{Illustration of the reasoning steps used by {\sc{KnowRob}} \citep{tenorth2009knowrob,tenorth2013knowrob} for inferring the location (i.e. in fridge) of a specific object (i.e. pancake mix) needed for cooking.  
    The semantic graph shows relations and properties about the objects and the current location.
    }
	\label{fig:roboearth}
\end{figure*}

When it comes to comprehensive knowledge representations that are based over the cloud, prominent examples include RoboEarth and RoboBrain.
RoboEarth \citep{waibel2011roboearth,riazuelo2015roboearth}, referred to as the ``World Wide Web for robots'', is an ongoing collaborative project aiming to create a cloud-based database for robots to access knowledge needed to solve a task.
RoboEarth was first proposed as a proof-of-concept to show that cloud robotics would greatly simplify robot learning.
RoboEarth provides an ontology for storage of semantic concepts and a method for accessing knowledge through a software as a service (SaaS) interface, where computational inferences and mappings can be done remotely.
As a collaborative system, RoboEarth allows robots to archive its own experiences (such as object types observed, motion plans successfully or unsuccessfully used, robot architectures, etc.) for recall and reuse by other capable robots.
This database would contain massive amounts of data (in the form of object models, SLAM maps \citep{riazuelo2014c2tam}, semantic maps, etc.) which can be used for tasks such as object recognition, instance segmentation and path-planning.
An illustration of the RoboEarth pipeline is shown as Figure \ref{fig:roboearth}.
Related to RoboEarth is another promising project headed by Rapyuta Robotics\footnote{Rapyuta Robotics - \url{https://www.rapyuta-robotics.com/}}, a company which now deals with cloud robotics solutions.
Rapyuta Robotic's system called {\it Rapyuta}, named after the movie from Japan's Studio Ghibli, was first introduced in \citep{hunziker2013rapyuta} and then in \citep{mohanarajah2015rapyuta} as a platform as a service (PaaS) interface for robots.
It acts as an open-source middleware for accessing resources from the web such as the aforementioned RoboEarth repository and ROS (Robot Operating System) packages.
Additionally, it reduces the processing done by the robot by offloading computations to the Cloud.
Robots can also communicate and share information with one another through this PaaS system.
This project has since evolved into the development of their cloud robotics platform for corporate solutions.

Results from RoboEarth led into the development of {\sc{openEASE}}\footnote{{\sc{openEASE}} - \url{http://www.open-ease.org/}}, also by Beetz et al. \citep{beetz2015open} (EASE being an abbreviation for {\it Everyday Activity Science and Engineering}). builds upon RoboEarth as a web-based interface and processing service that equips robots with knowledge from prior experiences (similar to accessing memory) and reasoning capabilities in the form of semantically labelled activity data. 
A robot using {\sc{openEASE}} will have access to: 1) knowledge about a robot's hardware, its capabilities, its environment and objects it can manipulate, 2) memorized experiences which a robot can use for reasoning (why it did an action, how it did it, and what effects the action caused), 3) annotated knowledge obtained from human demonstrations.
Queries and statements are formulated using ProLog, which can be sent through a web-based graphical interface or through a web API usable by robots; they allow robots to acquire semantic information and meaning to sensor input and to data structures used for control purposes.
As a component to this project, Tenorth et. al \citep{tenorth2009knowrob,tenorth2017representations,beetz2018know} presented {\sc{KnowRob}} (illustrated in Figure \ref{fig:knowrob}) as a knowledge processing system for querying the {\sc{openEASE}} knowledge base using Prolog predicates.
{\sc{KnowRob}} combines various sources of knowledge such as web pages (methods from instructional websites, images of usable objects, etc.), natural language tasks, and human observations.
A robot can ground the knowledge from {\sc{KnowRob}} to a robot's perception/action system and its internal data structures through a symbolic layer referred to as ``virtual knowledge bases''.
Through ProbCog \citep{jain2009}, a statistical relational learning and reasoning system, models such as Bayesian Logic Networks \citep{jain2013bayesian} or Markov Logic Networks can be built for representing the state of the robot's current context.
{\sc{KnowRob}} is built within another tool known as CRAM (short for {\it Cognitive Robot Abstract Machine}) \citep{beetz2010cram,beetz2012cognition}, a software toolbox for the design, implementation and deployment of robots using its own CRAM Plan Language (CPL). 
CPL is inspired by Common Lisp and Prolog for the expressive specification of concurrent, sensor-guided reactive behaviour, or in simpler terms, how a robot should react to certain sensory events or changes in belief state.

Another noteworthy technology that deals with knowledge gathering and sharing is RoboBrain\footnote{RoboBrain - \url{http://robobrain.me}}; 
Saxena et al. \citep{saxena2014robobrain} introduced RoboBrain in 2014 as a means of massively collecting concepts which are learned from automatic gathering of data from the Internet, simulations, and robot experiments.
This differs to the RoboEarth/{\sc{openEASE}} representation in the fact that RoboBrain uses graphs for encoding knowledge, while RoboEarth and its internal components use propositional logic and statements for defining concepts and relations in a working space.
The information is represented as a graph, where nodes represent concepts (such as images, text, videos, haptics data, affordances, deeply-learned features, etc.) and edges represe thent relatiohipns between such concepts.
RoboBrain connects knowledge from popular sources such as WordNet \citep{WordNet},  Wikipedia, Freebase, and ImageNet \citep{imagenet_cvpr09}. 
They manage errors in knowledge collection using crowd-sourcing feedback as well as beliefs that reflect the trust given to certain knowledge sources and the correctness of concepts and their relations.
To retrieve knowledge, the Robot Query Language (RQL) can be used for obtaining a subgraph describing the activity of executing a certain task. 
Unlike the case with {\sc{openEASE}}, it was not demonstrated how a robot can execute the method reflected by a subgraph; however, the knowledge gathered nevertheless can be quite useful for task planning, instance identification, and inference.

\begin{figure}[t]
	\centering
	\includegraphics[width=0.95\columnwidth]{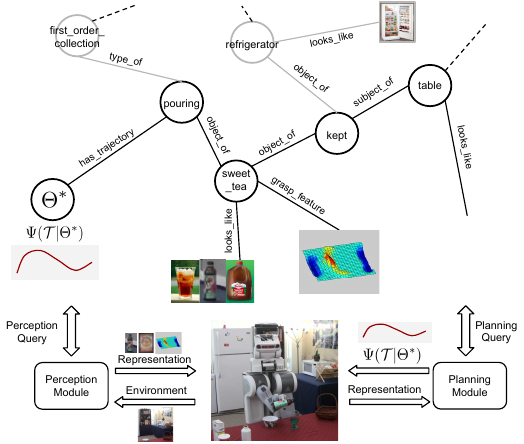}
	\caption{An illustration of how RoboBrain can be used by a robot for solving problems as originally shown in \citep{saxena2014robobrain}.
    }
	\label{fig:robobrain}
\end{figure}

\subsubsection{Remarks on Cloud-Based Systems}
To conclude, we have covered several knowledge representations that take different approaches to obtaining and retaining knowledge for robotics.
Although they all have their differences, the common goal is to develop a representation that is not specific to a single robot.
In fact, there are several ongoing projects whose aim is to take advantage of cloud computing platforms and technologies, and the abstraction of knowledge would therefore be essential.
Using distributed resources will also facilitate easier programming of robots through experience sharing and communication protocols that allow robots to work collaboratively.
However, the challenge lies in developing a generalized platform that can be used by different robots; since these robots are being built by a variety of companies and groups, it is up to them to foster seamless integration with these cloud services.
Close communication among researchers is crucial to ensure that research challenges are solved.
Additionally, defining and adopting universal standards or protocols would be key for the success of comprehensive and distributed representations.

\section{Specific Knowledge Representations}
\label{sec:models}
In contrast to full-fledged knowledge representations, there are other tools that can be useful to a robot in its execution of human-like manipulations and activities.
To create the ideal knowledge representation, such a representation must be comprehensive enough to tie high-level knowledge to low-level features/attributes; these models can be put together to integrate their strengths and to create a representation suitable for service robotics.
Researchers have extensively studied several types of models that can be used for representing different types of knowledge. 
Such types of knowledge are not limited to visual cues for grasp points or activity recognition, object-action-effect relationships to relate objects to their uses, object detectors and instance classifiers, motion identifiers, and object-object relationships.
The papers selected in this section consider these kinds of relationships when building their models. 

\subsection{Probabilistic Graphical Models}
In this section, we focus on learning approaches that use {\it probabilistic models} as their base of knowledge, which assume that a robot's world and the actions it can possibly execute are not discrete but indeterminate by nature.
In other words, the robot's world is governed by likelihoods and uncertainty, and these likelihoods are captured as probabilities grounded in such models.
These models therefore can be used for representing knowledge needed by robots when it comes to recognizing activities through a basal understanding of the effects of its own actions on its environment.
Although these models are examples of machine learning algorithms, they are fit to learn high-level concepts and rules for inference; other machine learning techniques that do not focus on these high-level rules would be considered as implicit representations of those rules, which will be covered in Section \ref{sec:ML}.
We first begin by covering the different types of models used in recent works, and then we explore how these models can be applied to learning and representing activities and actions.
We recommend that readers refer to works by Koller et al. \citep{koller2007graphical,koller2009probabilistic} for further reading on the theory of these probabilistic models.

\subsubsection{Bayesian Networks}
\begin{figure}[t]
	\centering
	\includegraphics[width=5cm]{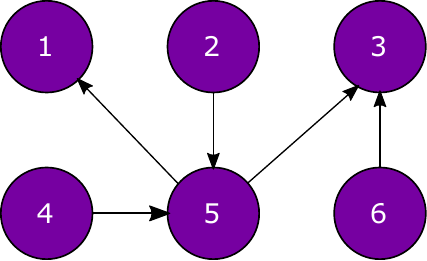}
	\caption{A simple example of a Bayesian Network.}
	\label{fig:BN}
\end{figure}

{\it Bayesian Networks} (BN) are directed, acyclic graphical models whose nodes can represent entities or concepts and edges denote probabilistic dependencies between such nodes and, furthermore, a flow of influence that may be incidental between them.
Each node in the network represents a conditional probability distribution, where probabilities are grounded by Bayes' Rule.
If we have an unknown variable, given that we know the other variables, we can use this equation to find the unknown probability or variable.
These variables are usually discrete and reflect observed phenomena within a setting or environment when applied to robotics; nodes are typically used to represent observable variables for properties or objects.
In simple terms, once there is an instance of a path between one variable to another, we can then say that there is a flow of influence since one node is dependent to the other; however, if there is no path connecting two nodes, such as the nodes labelled 1 and 2 in Figure \ref{fig:BN}, then we can say that these nodes are independent.
In mathematical notation, given that these node variables are $N_{1}$ and $N_{2}$ with no parents (hence, we say $\varnothing$), this is written as: \( (N_{1}\, \perp N_{2})\, |\, \varnothing \).
More accurately, this independence is best regarded as {\it conditional independence}.
Conditional independence assumes that two or more nodes follow a local Markov assumption, where a variable is said to be conditionally independent from nodes which are not its children given its parent node.
The presence of edges between a pair of variables indicate a flow of independence, while the absence of edges between a similar pair of variables indicate that they are completely independent.
It is because of this flow of influence that BNs are well-suited for inference about unobserved variables given those which we may have observed.

\subsubsection{Markov Networks}
\begin{figure}[t]
	\centering
	\includegraphics[width=6cm]{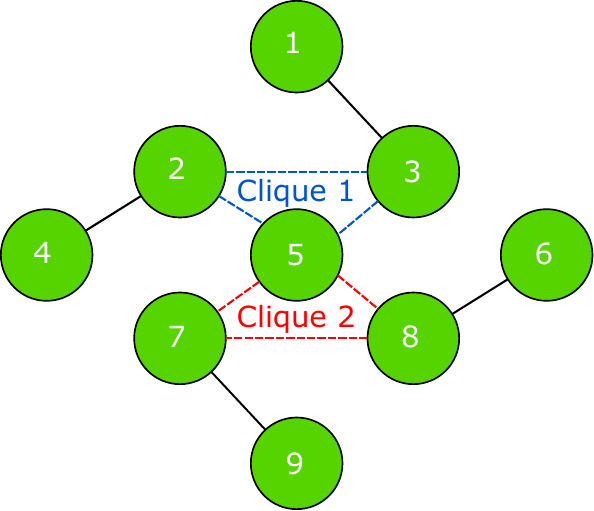}
	\caption{A simple example of a Markov Network.}
	\label{fig:MN}
\end{figure}

Bayesian Networks differ to another category of graphical methods referred to as {\it Markov Random Fields} (MRF) or {\it Markov Networks} (MN) with the stark difference in the graph's edges: there are no directed edges present in these structures. 
Since there is no directional flow of influence, a MN can adequately capture cyclic dependencies.
This means that there is an equal flow of influence by the pair of connected nodes.
As with BNs, we can also apply the local Markov assumption to denote conditional independence.
However, each variable in a MRF do not necessarily follow probability distributions. 
Instead, we parameterize a Markov Network using {\it factors}, which are functions representing the relationship for each {\it clique} (i.e. a subset of nodes in which all nodes are connected to one another - such as that in Figure \ref{fig:MN} represented by red and blue dashes) in the graph.
These factors, when combined together, can represent the entire state of the graph and make up a distribution.
These models can be used for representing the phenomenon reflected by given data just as with their directed variants, especially for representing transitions of states which are not necessarily irreversible.

\begin{figure}[t]
	\centering
	\includegraphics[width=7.5cm]{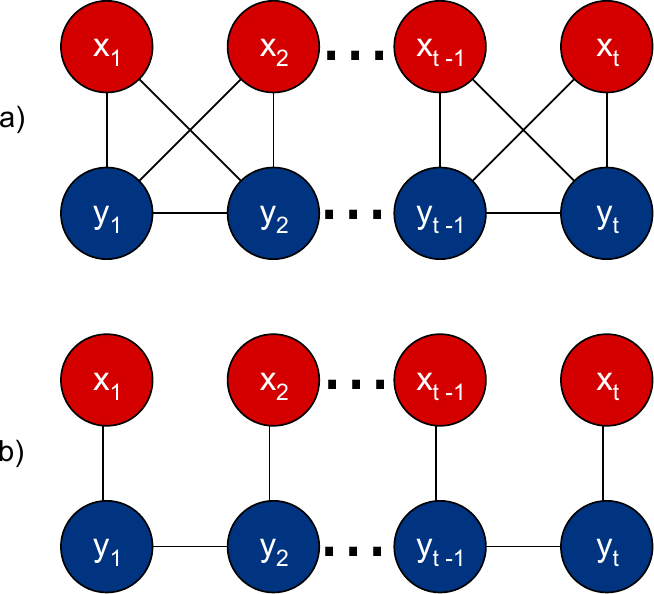}
	\caption{An example of conditional random fields.
    The CRF labelled a) and b) differ in terms of the states' dependencies on the observed variables. 
    CRF a) illustrates a model which uses several measurements for predicting a given state and this may be used for recognition, while CRF b) shows a simpler model which would only use the observation at some time $t$ to predict a state at time $t$. }
	\label{fig:CRF}
\end{figure}

\begin{figure}[t]
	\centering
	\includegraphics[width=5.2cm]{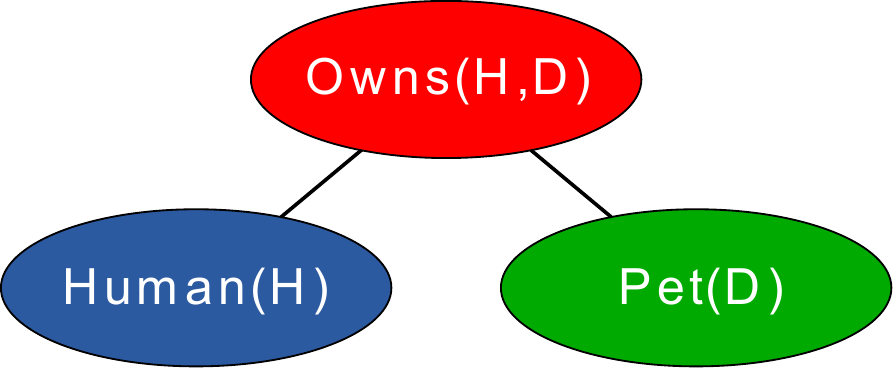}
	\caption{An example of a Markov Logic Network.
    This MLN represents the first-order logic statement that a human H can own a pet D (or more formally, \( \forall H, D : Owns(H,D) \rightarrow Human(H) \longleftrightarrow Pet(D) \)).
    For example, if we have instances of a human "Ryan" who owns a dog "Wilfred", then this would follow this logical statement where $Owns(\text{Ryan},\text{Wilfred})$, $Human(\text{Ryan})$ and $Pet(\text{Wilfred})$.}
	\label{fig:MLN}
\end{figure}

A Markov Network can be specialized further for specific purposes and functions.
{\it Conditional random fields} (CRF) \citep{lafferty2001conditional}, another instance of a Markovian model, are a special instance of MNs that encode conditional distributions (shown in Figure \ref{fig:CRF}).
Given a set of observed variables $X$ (representing data or measurements) and target variables $Y$ (representing states), a CRF models a distribution based on $X$, i.e. $P(Y|X)$. 
The definition of the distribution for a CRF would not solely consider factors (or cliques) with those observable variables (i.e. a given $C_i$ must not comprise of only variables in $X$).
CRFs are typically used for modelling predictions of sequential data, where the observable variables represent concepts that are used for predicting a certain outcome as target variables.
A {\it Markov Logic Network} (MLN) \citep{Richardson2006} uses the structure of Markov models combined with first-order logical statements or predicates, which describe the state of the world and those of objects and entities that are present within it.
These logical statements consist of four parts: constants, variables, functions, and predicates.
Each node is labelled with first-order logical statements to describe a probabilistic scenario, and they are assigned weights which reflect the likelihood of them being active.
A very simple example of a MLN based on a single first-order logic statement is shown in Figure \ref{fig:MLN}.
This differs from the regular usage of first-order statements: if a statement or condition is not true, then in the MLN, it means that it has a low probability of being true in that situation instead of being assigned zero probability.

\subsubsection{Applications of Probabilistic Models}
In this section, we discuss how these probabilistic models have been used for learning about activities and manipulations. 
Bayesian Networks in research studies are mainly used for capturing the effects of actions upon objects in a robot's environment.
When capturing such effects, a robot would be presented with demonstrations of observable action and effect pairs in order to learn the relationships between them.
These relationships can be taught through learning by demonstration, and it can be classified into two subcategories: {\it trajectory-level} learning and {\it symbolic-level} learning \citep{billard2008robot}.
Trajectory-level learning is a low-level representation approach which aims to generalize motions on the level of trajectories and to encode motions in joint, torque or task space, while symbolic-level learning looks at a high-level representation which focuses on the meaning behind skills in activity learning.
The robot's interaction with its environment serves to either learn new motor primitives or skills  (trajectory-level) or to learn new properties associated with the type of grasp they make or the skills they use, the object's physical features, and the effects that occur from executing an action (symbolic-level).
In works such as \citep{montesano2007modeling,montesano2008learning} \citep{moldovan2013use,moldovan2014learning,moldovan2017relational} \citep{stramandinoli2017heteroscedastic}, a robot can use basic, pre-programmed motor skills (viz. grasping, tapping or touching) to learn about relationships between an object's features (such as shapes, sizes or textures) and features of its actions (such as velocities and point-of-contact). 
The controllers of these skills are tuned by the robot's experiences and exploration with its environment, and the causality of these actions and their effects upon the world, based on object features, can be represented through a BN. 
The robot can use the knowledge it has gathered from interacting with objects and performing fine-tuning to select the appropriate action that achieves a human demonstrator's result.
Similarly, Jain et al. \citep{jain2013bayesian} and Stoytchev et al. \citep{stoytchev2005behavior,sinapov2007learning} used these networks to learn about the effects of actions on objects based on demonstrations with tools.
Their BNs were built based on geometric features relevant to a tool's function (and tools similar to it), which they coined as functional features, for predicting the effects of tools unknown to the robot with the learned model.
For instance, objects used for cutting have a sharp edge as a functional feature, and those used as containers have a non-convex shape for holding matter or substances; once the robot can identify these features, it can use the trained model to predict the results of specific actions.
The tools' effects are given by the target object's displacement, the initial position of the tool relative to the target object, and the target velocity after impact was made on the tool's functional feature.
A BN can also be used with other modalities of data such as speech input for grounding actions to their effects \citep{krunic2009affordance}.
Instead of learning object-action effects, BNs can also describe the likelihoods of object-object interaction between specific pairs of objects as learned from observing human behaviour \citep{SunRAS2013}.
These affordance models are particularly useful in improving both activity recognition and object classification and teaching robots to perform tasks using paired objects.

\begin{figure}[t]
	\centering
	\includegraphics[width=6.5cm]{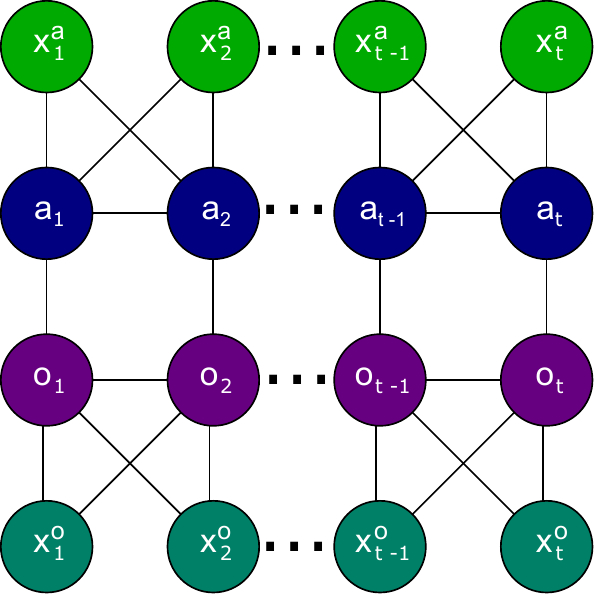}
	\caption{Illustration of a factorial CRF as used in Kjellstr\"{o}m et al. \citep{Kjellstrom} to predict actions $a$ and objects $o$ based on observed features $x^a$ and $x^o$ (for both action and object) at some given time $t$.}
	\label{fig:fCRF}
\end{figure}

With regard to Markov Networks, instead of a cause-effect relationship as inherently represented in Bayesian Networks, researchers can focus on learning dependencies between concepts useful for learning from demonstrations and identifying future cases of actions or activities, which can particularly be useful for a robot in learning new concepts or safely coordinating with others working in the scene.
As an example of activity recognition with MNs, Kjellstr\"{o}m et al. \citep{kjellstrom2008simultaneous,Kjellstrom} used CRFs to perform both object classification and action recognition for hand manipulations. 
Their reasoning behind the simultaneous classification of both objects and manipulations comes from: 
	1) the sequence of an object's viewpoints and occlusion from the hand indicate the type of action 
	taking place, and 
    2) the object's features suggest the way the hand will be shaped to grasp it.
They use factorial conditional random fields (FCRF) \citep{sutton2007dynamic} to map this two-way relationship between object features and possible action types (shown in Figure \ref{fig:fCRF}).
FCRFs have the advantage of mapping the relationship between the data level (features found in observations) and the label level (object types and properties and their relatedness with actions), thus effectively capturing affordances suggested by the hands and objects.
A similar approach is taken in \citep{pieropan2014} to identify activities based on objects and their proximity to hands in a scene using CRFs.
Prior to this work, in \citep{pieropan2013functional}, this association was described using text descriptions, which they denote as functional object string descriptors, which significantly perform better than using purely appearance-based descriptors for recognizing similar events or activities.
Using CRFs to represent the spatio-temporal relationship between objects, denoted by functional classes, improved over their previous approach of performing activity recognition with string kernels.

Another example of spatio-temporal representation of activities to objects was done by Koppula et al. \citep{koppula2013learning}, who used MRFs to describe the relationships present in the scene between activities and objects.
By segmenting the video to the point of obtaining sub-activity events, they can extract a MRF with nodes representing objects and the sub-activities they are observed in and edges representing:
	1) affordance-sub-activity relations (i.e. where the object's affordance depends on the sub-activity it is involved in),
	2) affordance-affordance relation (i.e. where one can infer the affordance(s) of a single object based on the affordances of objects around them),
	3) sub-activity change over time (i.e. the flow of sub-activities which make up a single activity), and
	4) affordance change over time (i.e. object affordances can change in time depending on the sub-activities they are involved in).
They proposed using this model for the purpose of recognizing full-body activities occurring in videos collected for their Cornell Activities Dataset\footnote{CDA - \url{http://pr.cs.cornell.edu/humanactivities/}} (CDA).
Following \citep{koppula2013learning}, they investigated how they can anticipate or predict human behaviour using CRFs to ensure that a robot reacts safely in \citep{koppula2016anticipating}.
A special CRF, called the anticipatory temporal CRF (ATCRF), can be built after identifying object affordance for a particular action type and can effectively describe all possible trajectories of human motion and sub-activities that are likely to be taken as time goes by.

Using first-order logic statements are effective for reasoning and inference; taking advantage of such logical expressions, a MLN effectively represents a systematic encoding of the robot's world.
A MLN can be thought of as a knowledge base, as these logical statements can be used for reasoning and drawing conclusions based on what a robot sees and observes.
For instance, with regards to activity recognition using affordance, Zhu et al. \citep{zhu2014reasoning} used a knowledge base, in the form of a Markov Logic Network, for inferring object affordances in activities which are suggested by the pose of the human demonstrator in videos.
They can do the inverse by predicting the objects and actions occurring in a given scene based on the pose of the human demonstrators with relatively great performance.
This could only be done after they collected a large amount of information about these usable objects and affordances as features, but there is no need for training multiple classifiers for each object-based task to identify each type of detectable activities as typically done with other machine learning approaches.
A MLN such as theirs can be used alongside other components for activity recognition to predict human intention and to enforce safe behaviour within a human-robot collaborative environment.
KnowLang \citep{vassev2012knowledge,vassev2015knowlang}, proposed by Vassev et al., is a knowledge representation that was developed for cognitive robots where the power of AI's logical expressions of the world with what they actually perceive in their world.
It also combines first-order logic with probabilistic methods which they can use for defining explicit knowledge for the robot.
However, when making certain decisions in which lies uncertainty, statistical reasoning through the use of Bayesian Networks makes the process more reliable through reasoning on beliefs.
Experiences can be reflected through probabilities, and such distributions are likely to change based on what the robot sees or acts.

\subsection{Semantic Graphs}

\begin{figure}[t]
	\centering
	\includegraphics[width=\columnwidth]{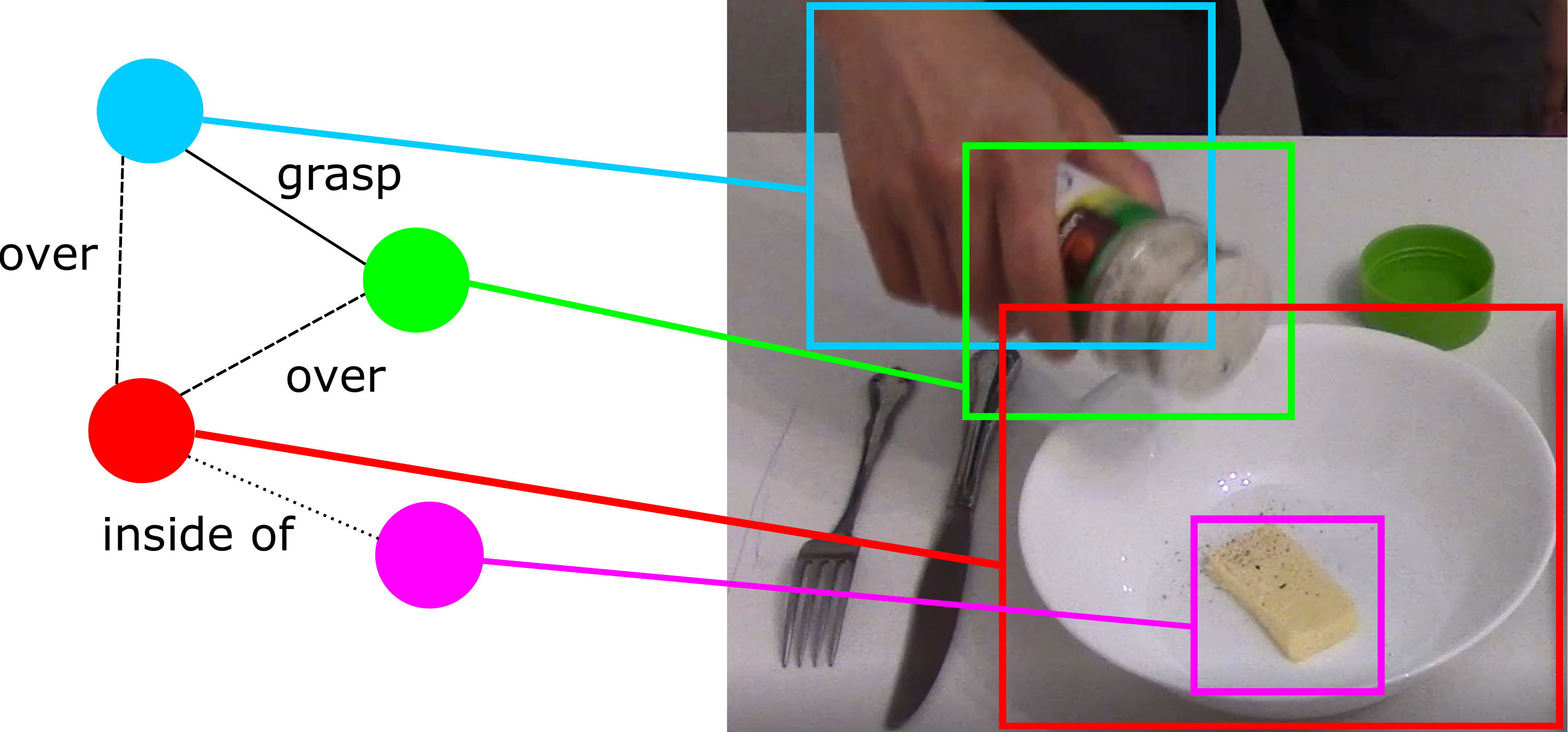}
	\caption{Illustration of semantic understanding of a certain scenario.
    In this case, we are creating a semantic graph whose nodes and edges describe the objects in the scene and their relative positions.}
	\label{fig:sem}
\end{figure}

Graphs are very popular for representing information quite simply because we can display knowledge in a graphical form that can be interpreted and verified visually by humans.
As previously discussed, probabilistic models can also be represented graphically and make excellent inference tools.
With probabilistic graphical models, edges would describe the likelihoods of certain variables as nodes causing others to occur.
However, in this section, we will be referring to another subset of graphs referred to as {\it semantic graphs}, whose nodes and edges describe semantic concepts and details between entities as observed in demonstrations.
Spatial concepts, for instance, can be described by semantic graphs, where nodes can describe objects within a scene, and edges describe commonality or contextual relationships between objects in terms of position (one object may hold another object, one object may be on top of another object, et cetera); such an example is shown as Figure \ref{fig:sem}.
Some graphs also embody temporal relations, where two or more particular events are related by time (e.g. one event must occur before another).
Basically, these networks can be used for compressing details and capturing relationships as needed by a robot.

\subsubsection{Activity Recognition \& Inference with Graphs}
One of the major problems in robot learning has been in learning to recognize activities to facilitate the transfer of knowledge to robotic systems.
A major component of activity recognition and understanding is predicting an ongoing activity or action as seen in a video demonstration.
Knowledge extraction is mainly done through the processing of activity-based videos and images or through interpreting sensor data from demonstrations either done by the robot or human demonstrators.
Techniques such as optical flow can also be used for identifying motion patterns to then recognize motion types or primitives. 
These elements can be used as context clues for inference.
Previous work focused on understanding the activity taking place with the use of the primary portion of such videos to recognize the likely activity and results which would be implied by it \citep{ryoo2011human,ryoo2015robot,cao2013recognize}, especially for predicting the next action which would take place in a sequence of actions \citep{vondrick2015anticipating}.
Semantic graphs have been used for representing affordances based on how objects are used with one another based on visual cues and spatio-temporal relatedness.

Segmentation techniques can be applied to images to identify the objects being used in demonstrations.
These segmented "patches" can be used for labelling nodes in semantic graphs (such as in Figure \ref{fig:sem}).
For example, Aksoy et al. \citep{aksoy2010categorizing,aksoy2011learning} created these semantic graphs after segmentation.
Their focus was in understanding the relationship between objects and hands in the environment and generalizing graphs for representing activities and identifying future instances of these events.
This approach can be classified as unsupervised learning since there is no explicit indication of what the objects are; objects instead are solely encoded based on manipulations in matrices, which they refer to as {\it semantic event chains} (SEC).
These structures capture the transitions in segment relations (temporal information), which are then generalized by removing any redundancies in activities, to be used in recognizing similar events.
They characterized spatial relations of objects as non-touching, overlapping, touching, or absent within each matrix entry and as edges which connect image segments.
Sridhar et al. \citep{sridhar2008learning} also used segmentation to separate the scene into "blobs" (similar to the patches in \citep{aksoy2010categorizing,aksoy2011learning} and cluster them as a semantic graph, based on the objects' usage in videos, for affordance detection.
Their semantic graphs are called {\it activity graphs}, structures which describe the spatial (whether objects are disconnected, found in the surrounding area, or touching) and temporal (relativity with respect to episodic events) relationships in a single video.
With such graphs, similarity between activities can be measured even with varying object instances, orientations, hand positions, and trajectories.
Zhu et al. \citep{zhu2015understanding} focused on segmenting the tool and the object it is being used on to create a spatial-temporal {\it parse graph}.
Within these graphs, they capture the pose taken by a demonstrator, the observed grasping point of the tool, the functional area of the tool that affords an action, the trajectory of motion, and the physical properties (such as force, pressure or speed) that govern the action.
These graphs can then be used to infer how objects can be used based on prior demonstrations.

\subsubsection{Representing Sequences of Skills or Events for Tasks}
Semantic graphs may also been used for task execution in the form of skills, containing knowledge that can be used by robots for manipulations.
In these structures, nodes represent objects and action types.
Several researchers have taken approaches to learning object affordance and representing them in this manner.
As discussed before, notable examples of such graphs include \citep{ramirez2015understanding,ramirez2017transferring} and \citep{Paulius2016,Paulius2018}, the latter being inspired by Petri Networks (or simply Petri Nets) \citep{Petri:2008}.
Petri Nets were originally intended for illustrating chemical reactions, and they have been shown to be applicable to other domains such as robotics and assembly.
Petri Nets are networks with two types of nodes: {\it place} nodes and {\it transition} nodes. 
Place nodes represent states of objects or entities, and transition nodes represent events or actions which cause a change in state.
The term for state change with respect to Petri Nets is {\it firing} of transitions.
Typically, all place nodes must be present for transitions to fire, therefore enforcing an implicit ordering of actions and behaviours.
Costelha et al. \citep{costelha2007modelling,costelha2012robot} used Petri Nets for representing robot tasks over other methods such as Finite State Automata (FSA) which require more memory and a larger space of representation and its limitation to single-robot systems.
Petri Nets, on the other hand, can represent concurrent system actions and sharing of resources.
The implicit ordering of events allows them to filter out specific plans which can never happen or those which should be avoided.
They created {\it Petri Net Plans} (PNP), which are essentially a combination of ordinary actions and sensing actions using control operators.

Similar to context-free grammars are {\it object-action complexes} (OAC, pronounced like ``oak") \citep{geib2006object,petrick2008representation,kruger2011object,wachter2013action}.
This representation's purpose is to capture changes in state of the environment in a formal structure which can be used for task execution.
OACs combine high-level planning and representation of the world with low-level robot control mechanisms called instantiated state transition fragment (ISTF).
An ISTF can be seen as minute, lower-level constructs, which can be put together like context-free grammars, for a concrete understanding of an action's effects before and after a motor program (i.e. action) is executed; OACs can be created after learning a variety of ISTFs.
ISTFs are generalized to only contain the object-state changes which are relevant to an action tuple (identified through methods described in \citep{aarno2007early}), as ISTFs can contain object-states which may or may not be affected or caused by a specific action.
Given a set of object affordances and relations learned, an associative network can be used for encoding and retrieving the state change that will occur from a certain OAC permutation.

Other approaches so far have attempted to map high-level manipulation skills to graphical structures.
Instead of focusing on manipulations, Konidaris et al. \citep{Konidaris} chose a different representation for trajectories as skills.
These researchers introduced an algorithm for learning skills from demonstrations, focusing primarily on motion trajectories from tasks, called CST (for {\it Constructing Skill Trees}).
Motion trajectories can be broken down using change-point detection, and these smaller trajectory components are referred to as skills. 
The aim of change-point detection is to find the point(s) at which there is a significant or observable change in trajectory.
After successfully segmenting the trajectories into skills, these learned skills can be combined together as {\it skill trees} for potentially novel manipulations by appending skills into one executable skill.

\begin{figure}[t]
	\centering
	\includegraphics[width=0.9\columnwidth]{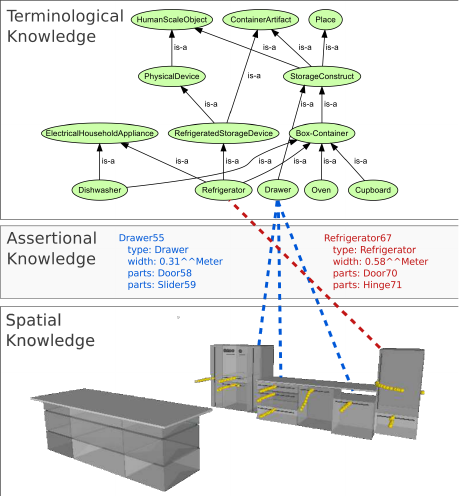}
	\caption{Illustration of various properties captured in the semantic object map (SOM) representation as presented in \citep{rusu2009model,pangercic2012semantic}. 
	SOM ties geometric data to semantic data that can be used in reasoning.}
	\label{fig:SOM}
\end{figure}

\begin{figure*}[t]
	\centering
	\includegraphics[width=16cm]{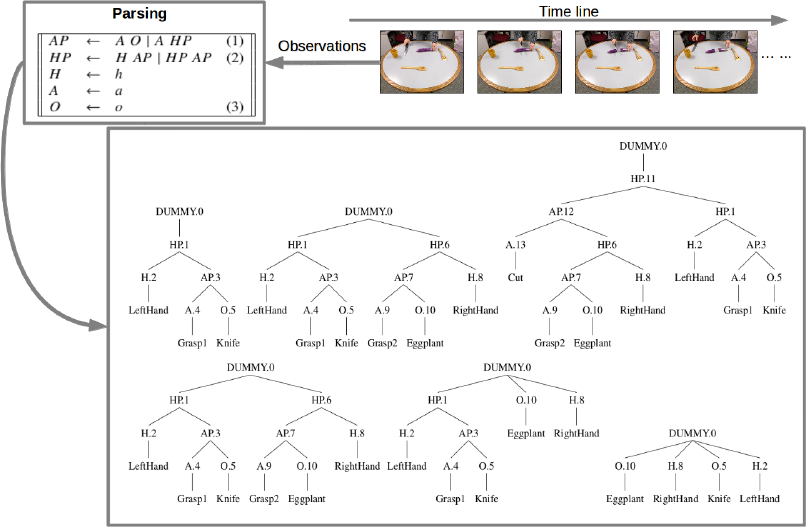}
	\caption{Illustration of how a context-free grammar can be used to construct parse trees as shown in \citep{YangAFA15}.
	The rules for their context-free grammar is described by production rules 1 to 3.
	Their manipulation action grammar includes the non-terminal symbols {\it A} describing the action, {\it O} describing the objects or tools being used within the activity, {\it H} describing the hand that is manipulating the object(s), {\it AP} and {\it HP} to describe the context of the action, hand and objects together.
	Actions based in this grammar can then used in robot task planning.}
	\label{fig:CFG}
\end{figure*}

\subsubsection{Combining Semantic and Physical Maps}
Semantic graphs can also take the form of {\it semantic maps}, which are special graphs that relate spatial or geometrical details (such as morphology of space, position and orientation of objects, geometry of objects as models, and any positions of special places of interest) to semantic descriptions (such as the purpose of objects).
These spatial features can be acquired from SLAM modules, including properties such as sensor readings or features, orientation, and absolute or relative positioning of objects or landmarks; through SLAM, the robot can obtain a map of the environment that uses the contextual information to particularly highlight instances of different objects or points of interest that lie there and to identify where they are, for instance.
Semantic maps have also been used in identifying grasps by using geometrical features about the objects \citep{dang2012semantic,dang2014semantic}.
An example of how semantic maps can be created was proposed by Galindo et al. \citep{galindo2008robot}, which integrates causal knowledge (how actions affect the state of the robot’s environment) and world knowledge (what the robot knows about objects around, their properties, and their relations), using two levels of information: the spatial box (S-Box) and terminological box (T-Box); they mark the physical location of objects at the sensor level as well as note the free space in rooms with S-Box, while the innate properties of these objects are linked using ontologies with T-Box.
{\it Semantic object maps} (SOM) \citep{rusu2009model,pangercic2012semantic} also serve a similar purpose to combine geometric data with semantic data to answer queries to determine whether a certain action can be executed given present circumstances in its environment.
For example, as in Figure \ref{fig:SOM}, a Room instance can be inferred to be a kitchen if there are items within the environment that are typical of a kitchen, such as a stove or a fridge.
With regards to creating semantic maps through human interaction, works such as \citep{bastianelli2013knowledge} \citep{kollar2013learning} and \citep{randelli2013knowledge} aimed to develop HRI systems to impart knowledge to a robot about its surroundings: what lies around it and the conceptual knowledge tied to these elements.
Both systems use audio for interacting with robots; in addition to this speech recognition system, \citep{randelli2013knowledge} combined a tangible user interface and a vision system with a robot's modules for motion planning and mapping to compile and create a knowledge base which a robot can then use for its navigation through its environment.

\subsection{Context-Free Grammars}
Context-free grammars are also an effective way of constructing or representing semantic graphs or structures, as they guarantee completeness, efficiency and correctness.
A context-free grammar defines rules to creating semantic structures and sub-structures as strings using its own symbols (called {\it terminals}) defined within a finite set called an {\it alphabet}. 
These terminal symbols can be used when substituting variable symbols called {\it non-terminals}; the substitution process is described by production rules, which allow us to generate sentences.
With such a formal definition of a context-free grammar, researchers have been able to define rules that describe concepts such as manipulation/action types which can then be useful for defining plans that robots can use for execution and also for the composition of skills into sub-skills.
One such example of a context-free grammar was proposed by Yang et al. \citep{Yang_2013_CVPR,yang2014cognitive,yang2014manipulation,yang2015grasp}, who studied how manipulations in activities can be represented through grammar and then broken down into visual semantic graphs.
These parse trees (as described in Figure \ref{fig:CFG}) are built from demonstrations can then be used to form manipulation action tree banks.
The high-level representation serves as a symbolic representation of the manipulations which describe each step required to solve a given problem.
In a different approach that also uses context-free grammars (CFGs), Dantam et al. \citep{dantam2011motion,dantam2013motion} also formulated robot primitives and control policies using their own representation called {\it Motion Grammars} (MG).
A parse tree can be constructed to reflect the procedures being executed and they can be broken down by a motion parser to create sub-tasks until they have been satisfied similar to the representation introduced by Yang et al.
The experiments conducted in their work showed that a robot can effectively execute game-related manipulations while maintaining a safe environment for humans playing with it.
Even though there was no discussion on how these can be used for service robotics manipulations such as domestic or industrial tasks, the use of MG can ensure safe, real-time control of robot tasks.

\section{Model-Level Representations}
\label{sec:ML}
The aim of machine learning algorithms is to find a suitable model or function for discriminating classes or for predicting effects or labels; techniques such as neural networks and probabilistic graphical methods can all be considered as machine learning algorithms in addition to those discussed within this section.
Although we have previously discussed some of these models, they were used in the context of learning high-level rules or inter-relationships between concepts (such as the object-action-effect relationship).
They are equally suitable along with other machine learning approaches to learn lower-level concepts that are not sufficient to explicitly represent knowledge and to perform reasoning tasks. 
However, when applied, classifiers can implicitly represent rules that describe classes needed for a robot's tasks.
Therefore, such learning models complement knowledge representations in performing specialized tasks, such as image understanding, segmentation and processing, to then further guide a robot's future actions.

\begin{figure}[t]
	\centering
	\includegraphics[width=0.95\columnwidth]{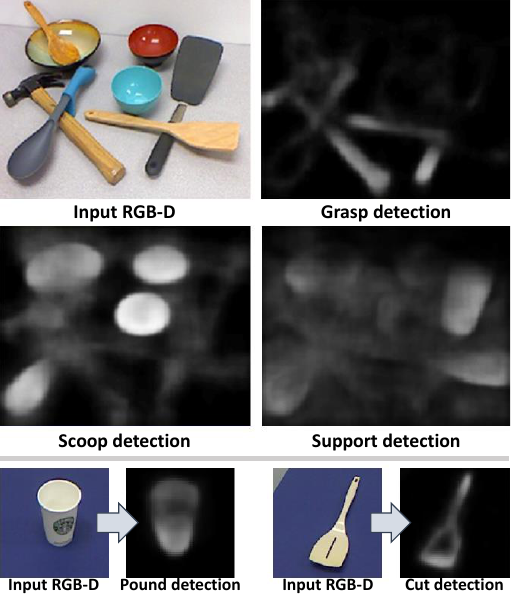}
	\caption{An example of affordance-based pixel hot spot detection as done by Myers et al. \citep{myers2015affordance}.
    	In their project, they can look at a scene and identify the types of operations a robot can perform using objects found there by highlighting tools that afford a specific action or purpose. A robot can use this to further decide its actions.}
	\label{fig:myers}
\end{figure}

\subsection{Classical Machine Learning Techniques}
Major successes in reinforcement and imitation learning can been attributed to machine learning methods \citep{peters2003reinforcement,kober2013reinforcement,deisenroth2013survey,Argall}, proving them to be useful for training robots in a bottom-up fashion.
Just as we have addressed before with other models such as probabilistic models, imitation learning aims to teach a robot about skills that can be incrementally learned and used for task planning.
Machine learning algorithms have been used in this regard to learn new motor primitives or skills \citep{peters2007machine,kober2009learning,aarno2007early,ugur2012self} or to successfully classify or predict the outcomes of actions on objects \citep{tikhanoff2013exploring} \citep{ugur2014bootstrapping,ugur2014emergent}.
Works like \citep{aarno2007early} and \citep{ugur2012self} apply machine learning algorithms to learn how to grasp objects based on their features.
In \citep{tikhanoff2013exploring}, the researchers aimed to represent the effects of the robot's motions on objects in its environment using a support vector machine (or SVM, a model which can be used to discriminate between classes by representing data in high dimensions and finding a function or hyperplane that optimally separates class instances), which, once trained, can be useful in predicting the change in an object's state that is reflected by its displacement in the scene when performing actions (e.g. whether the objects would roll or respond to pulling).
SVMs are also used in \citep{ugur2014bootstrapping,ugur2014emergent} to learn about object effects (such as push-ability and stack-ability) based on physical properties of objects. 
To better understand the consequences of its actions, they tweaked various parameters such as the hand speed and the tactile sensors in the robot's hand.
For the task of affordance detection, \citep{myers2015affordance} used two classifier variants, hierarchical mapping pursuit (or HMP, which are similar to neural networks) and random forests (an ensemble of decision trees or similar classifiers), to identify pixel hot spots within a scene that suggest features related to affordances; an illustration of this task is shown as Figure \ref{fig:myers}.

\subsection{Neural Networks for Deep Learning}
A vast number of studies have taken advantage of the effective method of {\it deep learning} and the use of neural networks for tasks such as handwritten character recognition \citep{lecun1989backpropagation,lecun1998gradient} and image processing and recognition \citep{krizhevsky2012imagenet,simonyan2014very,zeiler2014visualizing}.
The power in deep learning techniques lies in its property as a ``universal approximator" \citep{hornik1989multilayer}, where it can theoretically learn any model or function given the right number of layers (meaning one or more hidden layers).
A deep neural network represents a feature detector or classifier which can be stored and reused by processing systems for specific purposes such as identifying affordances, grasping points, object recognition, instance segmentation, and many more.
The amount of training time can be significantly reduced by using pre-trained models and fine-tuning them with other data sets, since the networks will retain valuable information that can be extended to other domains.
For instance, it is customary to use pre-trained versions of well-established networks such as VGG \citep{simonyan2014very}, ResNet \citep{he2016deep} and GoogLeNet \citep{szegedy2015going}, which have all been successfully applied to the task of labelling images.
We refrain from going too deep into this machine learning technique in this paper, and so we refer readers to \citep{schmidhuber2015deep,liu2017survey} for extensive reviews on deep learning and neural networks.

Neural networks have been proven to be capable of learning very effectively in both supervised and unsupervised learning for robotics.
As a component to a knowledge representation, these models can be used by a robot to handle decisions or disputes among a variety of choices that may not be easily grounded in explicit rules. 
One task for which they have particularly been effective for is grasp synthesis; in \citep{bohg2014data}, the authors review the research in data-driven grasp synthesis, where the ideal or optimal grasp type is selected among a sampling set of grasps.
A neural network (such as those in \citep{mahler2016dex,mahler2017dex,mahler2017dex3})
would be used to recommend the ideal point to grasp objects, even those that have never been seen before \citep{lenz2015deep,levine2016learning}.
In addition to robotic grasping, these networks are also effective for scene understanding \citep{chen2016deep,zhang2017deepcontext,aditya2018image}, specifically affordance detection within a scene \citep{nguyen2016detecting,srikantha2016weakly,AffordanceNet18}. 
They have also been applied to the generation of textual descriptions of what they observe in their environment \citep{8258893} for communication between robots and humans.
The identification of what is around the robot can also be referenced to underlying rules, facts, or beliefs.
Deep learning works well in activity recognition due to their exceptional performance in image segmentation, object recognition and instance identification \citep{simonyan2014very,gupta2014learning,long2015fully}, even in real-time \citep{dvornik2017blitznet}.
Such networks can be used for learning task plans from demonstrations to build upon a robot's knowledge base \citep{yang2015robot,sung2018robobarista}; these learned plans can be used for predicting future activities to anticipate the consequences and forces exerted from those actions \citep{fermuller2018prediction}.
Finally, neural networks can be used in learning control and the dynamics/physics of actions \citep{wu2015galileo,finn2016unsupervised,battaglia2016interaction,yildirim2017physical,byravan2017se3}.

\subsection{Summary}
To conclude, many of these works have shown how machine learning algorithms are effective in the classification of a wide array of applications such as detecting object affordances and grasp points.
Traditionally, algorithms such as SVMs have been used for a wide array of tasks and are now being replaced with deep learned networks since neural networks are easy to train.
Neural networks are especially very useful for extending applications to simpler systems, and they have proved to work exceptionally well in image processing tasks.
We can use trained networks to help us identify key features that can be adopted to a programmed approach or to other rule-based classifiers.
Despite their great success and trending usage among the robotics community, we should not consider them as the ``magic bullet" for all problems \citep{sunderhauf2018limits}.
They simply cannot serve as stand-alone knowledge representations due to the lack of explicit knowledge and meaning behind what they implicitly represent; nevertheless, they are quite useful for guiding the decision-making process for a robot.

\section{Selecting Knowledge Representations}
\label{sec:eval}
\begin{figure}[t]
	\centering
	\includegraphics[width=0.85\columnwidth]{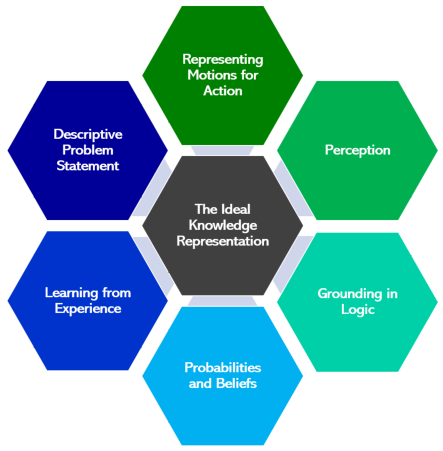}
	\caption{Overview of Section \ref{sec:eval}. When designing or selecting a knowledge representation, researchers should take into consideration as many of these concepts as possible;
	this however is a very challenging problem, as there is no fundamental way to integrate these components together into one representation.}
	\label{fig:eval}
\end{figure}

Previously, we examined several different approaches that have been implemented successfully to solve a wide array of sub-problems needed to build a complete and comprehensive knowledge representation.
Each technique comes with its own limitations and strengths, and they should be used accordingly with one another to create an ideal knowledge representation.
In this section, we would like to pay closer attention to the key features of knowledge representations as observed in the works discussed in this paper.
We argue that the following components (depicted as Figure \ref{fig:eval}) are needed for a sufficient knowledge representation:
    1) the representation of motions or skills for task planning at the semantic level of understanding;
    2) a perception system, which integrates multiple channels of input such as vision, haptics, natural language, audio or sensors to localize objects, tools, or entities (humans, animals, etc.), to perceive observable states as well as possible obstacles, and to communicate with other robots or humans;
    3) the grounding of perception or control to logical statements, symbols, or values, which can be used to connect concepts and solve problems through logical reasoning;
    4) the retention of experiences as beliefs to represent uncertainty, which can influence a robot to consider other possibilities in task planning as well as tailoring the robot's actions to the human beings or environment it has been built to operate under (through statistical reasoning);
    5) the committing of newly acquired knowledge and concepts for re-use as well as the ability to learn new concepts;
    6) a suitable definition and representation of the (household) environment.
A majority of these things are unfortunately not a focus in several of the representations discussed in this paper, as these components are usually built independently from others.
Perhaps the closest representation to this idea is {\sc{KnowRob}} through {\sc{openEASE}} \citep{tenorth2010knowledge}; nevertheless, these systems still require a lot of work to get them to the standard of working reliably among humans.
In the following subsections, we discuss these different ideas and why they are important for knowledge representations used in service robotics.

\subsection{Component \#1: Representing Motions for Action}
\label{sec:com-1}
Perhaps the most important of all components, a knowledge representation gives meaning to a robot's actions and environment; the way we represent these aspects are very crucial to building an effective knowledge representation.
In particular, we should focus on meaning behind a robot's actions and motions taken for the understanding the consequences of its actions.
A representation of actions, whether grounding them as motion primitives or denoting an action as a combination of skills, can be used for signifying meaning behind the transition between object states (or quite simply, the consequences of performing those actions).
When reasoning and determining its goal, the robot needs to find the right actions to get to that goal.
Saving these primitives as skills also facilitates re-use, and they can always be tuned to different parameters based on what the robot perceives.
The organization of activities or tasks as motion/skill trees makes it easy to combine skills or primitives according to the current problem and the present state of the robot's world.
Whether it be representative of trajectories as in \citep{Konidaris} or individual skills \citep{ramirez2014bootstrapping,ramirez2015understanding,ramirez2017transferring}, \citep{geib2006object,petrick2008representation,kruger2011object,wachter2013action}, \citep{yang2014manipulation}, the idea that ties them together is the ability to combine multiple units together to suit the problem at hand.
Using combinatorial representations such as \citep{Yang_2013_CVPR,yang2014cognitive,yang2014manipulation,yang2015grasp}, \citep{Konidaris} and \citep{dantam2011motion,dantam2013motion}) also emphasize the idea of compositionality and flexibility of structures for task planning.

\subsection{Component \#2: Perception}
\label{sec:com-2}
An important component to any intelligent agent is perception, as an agent such as a robot needs to acquire input to determine how to proceed with future actions.
For instance, a robot would need to acquire input to determine how it would trigger conditions for rules such as in Figure \ref{fig:Prolog}.
Many solutions discussed in this paper under Section \ref{sec:ML} have been shown to be effective in processing input to determine grasp points, object affordance hot-spots, object recognition and instance segmentation, etc.
Other classical computer vision techniques have also been used for processing input through images or videos to perform similar tasks.
A knowledge representation can also store useful object descriptors in forms such as 3D models or point clouds, which are made available (implicitly or explicitly) in works such as \citep{waibel2011roboearth,riazuelo2015roboearth},  \citep{saxena2014robobrain,sung2018robobarista} and \citep{mahler2016dex,mahler2017dex,mahler2017dex3}.
Now that we have shown that we can detect and process all manner of input, we should instead focus on how this can be done in {\it real-time}.
In addition to this, the models used for perception should be trained to handle uncertainty as best as they can by training on a wider set of samples for the corresponding household tasks the robot is given; this brings us to another point in selecting suitable training samples to give a general representation of several concepts to work with.
For instance, one possible limitation to machine learning models is the likelihood of overfitting to the training set, causing the model to develop a bias towards the data rather than learning what is really important for classification.

\subsection{Component \#3: Grounding Concepts with Logic}
\label{sec:com-3}
As a basic premise of AI, knowledge representations allow for reasoning using logical statements and expressions by using explicit knowledge to make inferences to acquire implicit knowledge describing the current state of the world that then reflects the next likely action to be taken.
Structures such as MLN (as used in \citep{zhu2014reasoning} and \citep{vassev2012knowledge,vassev2015knowlang}) integrate logical rules with probabilities.
However, many of the learning models that have been developed lack the grounding of such concepts to explicit statements.
Without doing so, it becomes a challenge to define expectations of actions, such as the change in the state of the robot's environment; when ignoring states, it becomes difficult for a robot to determine whether it has completed the action successfully or not \citep{jelodar2018identifying}, and we cannot always assume that the robot will solve problems without fail.
Rules that imply an order to actions and tasks make it easy to modularize actions.
For instance, in \citep{Paulius2016,Paulius2018,jelodar2018long}, states are explicitly reflected in object nodes within the network, and these can be used for indicating whether a robot has successfully performed the manipulation or not while also representing precedence of its actions.
Logical formulations of concepts also help us to translate human-understood concepts into a ``mid-level" knowledge that can be understood by both humans and robots alike.

\subsection{Component \#4: Probabilities \& Beliefs}
\label{sec:com-4}
Realistically, an agent cannot fully depend on logic to understand its world, as predictive modelling does not always give a straightforward answer.
It is through the use of beliefs that a robot can statistically reason to make decisions about its future actions.
Several classifiers, for instance, cannot give a 100\% accurate answer about what it is identifying, and so weights are assigned to give the best guess of what is presented to the robot.
Another example would be with a robot's sensors: innately, sensors capture noise from the input it gathers from its surroundings.
A robot should leverage its beliefs or estimates with measurements because it will not always have a completely correct understanding of its world.
The use of probabilities also makes for strong inferences through probabilistic models as discussed in prior sections.
Beliefs can also be updated to make stronger inferences about what a robot sees around itself to make smarter decisions.
For instance, by using information about object affordances, a robot can select the right tool for a particular job; identification of said tools can be done using learned classifiers such as \citep{nguyen2016detecting,srikantha2016weakly,AffordanceNet18,myers2015affordance}.
With respect to previously discussed works, probabilistic and statistical reasoning methods have been used in learning how its actions affect its environment.
Leveraging explicit knowledge using a robot's beliefs about its surroundings, through both logical and statistical reasoning, would allow it to make the best decision possible.
For example, in the task of SLAM, the joint task of location estimation and update of its belief and location is important for the robot to proceed with navigating its environment with as minimal error as possible.
Weights may also aid the robot's system in deciding between multiple courses of action so as to reduce the likelihood of task failure.

\subsection{Component \#5: Learning from Experiences}
\label{sec:com-5}
To guide a robot in performing its duties, it is beneficial for it to retain knowledge from prior experiences in solving a problem (or failing to do so).
This is a key advantage that is contained in knowledge representations such as {\sc{openEASE}}, RoboEarth and {\sc{KnowRob}}, where a robot can record details about the manipulations it has performed over a distributed system.
Other robots that are attempting to perform the same manipulation can then access those prior experiences to adapt its actions to its problem; details such as trajectories can help to guide the current robot to follow similar patterns.
More specifically, in the {\sc{KnowRob}} representation from Section \ref{sec:cloud}, one of the key aspects of the platform is to record the experiences of performing a specific task or manipulation.
As in their recent work in \citep{beetz2018know}, they illustrated how one robot can use the experience of opening a fridge from another robot to perform the same task; however, the exact trajectory as learned from the previous robot cannot be completely used due to differences in the state of the targeted object and its environment.
Therefore, it is not simply imitating what a previous robot has done, but rather, it refines the motion to better suit the current problem.
As another example of alternatives for future task planning, aside from others in Section \ref{sec:high-level}, the FOON representation draws from multiple demonstrations to potentially create novel manipulation sequences.
This is done through the merging of subgraphs for each demonstration into one universal FOON.
With machine learning models, the overall idea of training these models is to adapt them to accurately predict future instances.
Probabilistic models, for instance, have been used for predicting the effect of actions on objects, whether physically or semantically.
The ideal representation therefore should retain experiences to guide the robot's actions.

\subsection{Component \#6: Descriptive Problem Statement}
\label{sec:com-6}
Finally, the major difficulty in creating the ideal knowledge representation is the inability to define a robot's world precise enough to solve every possible problem.
As we talked about before, logical expressions can be useful for grounding causal rules, while probabilistic expressions can help in answering ambiguities using beliefs.
However, to vividly paint the picture of the robot's world in different structures and databases in this way really is no simple matter.
A clear definition of the problem would help to identify those areas in robot manipulation and learning that we have done well enough and those areas that we still have a long way to go with.
For instance, by clearly defining or constraining the environment in which a robot operates in, we can build tools or models that will function well even in variability.
The idea of building more robust classifiers is mentioned in \citep{scheirer2014probability} through the idea of {\it open set recognition} \citep{bendale2015towards,bendale2016towards,sunderhauf2018limits}, where the classification is redefined to account for robust detection of known positive classes while rejecting both known and unknown negative classes.
The formulation of what the robot needs to identify versus what does not concern it for a particular action therefore is very important to clearly define.

\subsection{Final Analysis}
Given that we keep the previously mentioned components in mind, we can develop more effective representations for robots.
It is important that we as researchers consider how we can develop a robot that understands its environment and its own actions to induce smarter behaviour.
Instead of confining a robot to a single environment by programming it in a ``hard-coded" fashion, through a properly designed knowledge representation, a robot can work in variants of the same environment it is trained to work among (for example, one robot trained to work in kitchens can recognize how a kitchen looks like and what it expects to find within a kitchen).
Humans can map their own commands and understanding in that terminology, because we do not understand the world in low-level representations as used in machine learning algorithms or numbers for colour values, depths, et cetera.
As in AI, a robot will be able to intelligently gather new knowledge and make its own decisions and actions based on its experiences (or those of other robots).
Several sub-tasks in robot problem solving can rely on machine learning models or tools to derive a solution, and we believe that these tools should adequately be used in unison with others.
However, it is still important to derive representations filled with meaning, as implicit representations of knowledge can lead to unpredictable results, and machine learning models are not solutions for intelligent behaviour in robots.
An ongoing issue is the development of a standard for representations and ontology \citep{prestes2013towards} and it is important to determine a common goal in the development of independent knowledge representations.
This would eventually lead to promising developments for the future of robotics to integrate multiple platforms and to sustain cross-architecture communication and collaboration between robots through a common language, interface and/or network.
In addition, we suggest that researchers focus more on building representations that can be built and accessed remotely through databases or cloud systems to make robot learning easier while reducing the strain on a robot's on-board system to facilitate working in real-time scenarios, such as RoboEarth and {\sc{openEASE}}.
The ability to reproduce actions or sequences and the retention of experiences as memory would make training robots more efficient and less time consuming, since we do not necessarily have to learn tasks or skills from scratch.

\section{Concluding Remarks}
To summarize, this paper provides an overview of technologies and research that is being used in robot learning and representation for service robots. 
To suitably define a knowledge representation, it is important to distinguish representation from learning.
Machine learning models, for instance, are not sufficient to be representations alone and are better referred to as specialized learning models or classifiers.
However, the effectiveness of such models should not be disregarded; rather, we should emphasize the \emph{integration} of multiple components together with high-level knowledge to fully equip a robot with the means to understand its environment and actions to perform meaningful tasks.
It is important to note that this paper has not talked about the efficiency of  representation and retrieval algorithms, but instead this paper tackles the principles behind an effective knowledge representation for a robot to use to its fullest potential.
Ideally, a knowledge representation should encompass all of the necessary information needed by a robot to solve a problem and understand its environment in order to do so effectively.
It should allow the robot to successfully perform several actions such as perception/vision, knowledge acquisition, motion and task planning, and reasoning.
However, due to the highly variable nature of domestic environments and households as well as the question of safe operation among humans, building a suitable comprehensive knowledge representation becomes a very challenging and complex problem. 
In order to build suitable knowledge representations, we should strive to break this larger problem into smaller problem sets; we should take advantage of the many models that solve several sub-problems that a robot would typically be confronted with in manipulation problems.
To further improve on the usefulness of knowledge representations, researchers can take advantage of large-scale distributed systems and networks for sharing knowledge and experiences that would help to teach robots how to do certain tasks from scratch without the need for reprogramming or rebuilding solutions.
These distributed representations, such as those discussed in Section \ref{sec:cloud} can be hosted and built upon through the cloud or on the Internet.
Although challenging and ambitious, a cloud computing interface for robots can greatly advance robotics technology.
The most important takeaway from this paper is that there is always a need for standards to be developed and adopted by researchers in the realm of service robotics in order for safe and efficient robots to be deployed among humans.

\section*{Acknowledgement}
We would like to acknowledge the National Science Foundation for supporting our work under Grants No. 1421418 and 1560761.
We would also like to thank our reviewers for their insightful and thorough comments that helped to greatly improve this manuscript.

\biboptions{square,comma}
\bibliographystyle{elsarticle-num}
\bibliography{ref}

\end{document}